# An Automated Multimodal Glaucoma Detection Framework Using ViT and a Stacking-Based Ensemble

**Authors:** Ishrat Jahan[1], Muhammad E.H Chowdhury[2], Murugappan Murugappan[3,*], Kanchon Kanti Podder[4], Tawsifur Rahman[5], Shrestha Datta[1], Md Sakib Abrar Hossain[6], Md Mosarrof Hossen[2], Yosra Magdi Salih Mekki [7], Sanjiban Sekhar Roy[8]

**Affiliation:**

1 Department of Computer Science and Engineering, Shahjalal University of Science and Technology, Sylhet, Bangladesh; i.jahan1533@gmail.com, shresthadatta910@gmail.com

2 Department of Electrical Engineering, Qatar University, Doha, Qatar; mchowdhury@qu.edu.qa

3 Department of Electronics and Communication Engineering, Kuwait College of Science and Technology, Block 4, Doha, Kuwait; m.murugappan@gmail.com

4 Department of Interdisciplinary Engineering, Kennesaw State University, United States; kk.podder754@gmail.com

5 Department of Biomedical Engineering, School of Medicine, Johns Hopkins University, United States; tawsifurrahman.1426@gmail.com

6 Department of Bioinformatics University of Regina, Canada; hossensakib27@gmail.com

7 Department of Biomedical Engineering, University of Oxford, United Kingdom; yosra.mekki@eng.ox.ac.uk

8 Department of Computer Science and Engineering, Vellore Institute of Technology, India; sanjibanroy09@gmail.com

**Abstract**

Glaucoma is a progressive eye disease that can lead to irreversible vision loss if not detected at an early stage. Conventional diagnostic procedures are often time-consuming and rely heavily on expert interpretation, limiting their scalability for large-scale screening. In this study, glaucoma detection is investigated under two evaluation settings: sample-wise, where individual samples are analyzed independently, and patient-wise, where data from each patient are aggregated for final prediction. An automated multimodal framework is proposed that integrates fundus images with clinical data. Under the sample-wise setting, detection is performed using fundus images and clinical features individually, as well as through their multimodal combination. Under the patient-wise setting, predictions are obtained by aggregating multiple fundus image representations with corresponding clinical information for each patient. Deep visual features are extracted using a Vision Transformer (ViT) architecture and classified using classical machine-learning models, with a stacking-based ensemble of the three best-performing classifiers employed to optimize performance. Experiments conducted on the publicly available PAPILA dataset demonstrate strong diagnostic performance, achieving 97.47% accuracy and a 97.50% F1-score for sample-wise multimodal classification, and 98.97% accuracy and F1-score for subject-wise detection. The proposed framework is further deployed as an end-to-end web-based platform to support automated glaucoma screening and clinical decision support.

## 1. Introduction

Glaucoma is the world's second-leading cause of blindness [1], affecting 4.5 million blind individuals globally [2]. Approximately 76 million people suffer from this disease, which worsens with age [3-5]. It affects approximately 2 to 3% of adults over 40 and up to 50% of cases may remain undiagnosed due to the late manifestation of symptoms [6, 7]. This condition leads to varied rates of blindness globally, with the lowest in South Asia and the highest in Africa [8,9]. Consequently, many patients seek treatment only when their vision deteriorates. Primary open-angle glaucoma (POAG), the most prevalent form, represents 90% of cases. Studies in 2021 reveal that a significant percentage of POAG patients had never been diagnosed [10]. Other types include closed-angle glaucoma, affecting 20 million people with 5 million sightless [11], and secondary glaucoma, which arises from other types of eye ailments or long-term use of certain drugs [12]. Congenital glaucoma, affecting one in ten thousand infants, often results in blindness due to delayed treatment [13].

The optic nerve transmits visual signals from the eye to the brain, making vision possible [14, 15]. The eye contains a clear fluid called aqueous humor, which helps nourish internal eye structures and maintain normal eye shape. This fluid is produced from plasma by the ciliary epithelium of the pars plicata through both passive and active secretion [16]. In addition to maintaining the optical clarity of the cornea, anterior chamber, and crystalline lens, aqueous humor plays an important role in regulating intraocular pressure (IOP) [17]. Normal IOP depends on a balance between the production and drainage of this fluid [18]. In glaucoma, this balance is disrupted, causing aqueous humor to accumulate and increase IOP. Elevated eye pressure or reduced blood flow can damage the optic nerve, leading to vision loss. Both intraocular and intracranial pressures are believed to contribute to glaucoma development [19]. Normal intraocular pressure typically ranges between 12 and 21 mm Hg, while values above 21 mm Hg are considered abnormal [20]. Sustained pressures in the range of 20–30 mm Hg can gradually damage ocular structures, whereas pressures of 40–50 mm Hg may lead to severe vision loss. Although damaged retinal neurons cannot be restored, medications can slow the progression of glaucoma [9]. The disease often reduces quality of life and increases both personal and healthcare-related costs [21]. In patients with elevated intraocular pressure ($\geq$21 mm Hg), lowering eye pressure by about 22.5% can reduce the risk of open-angle glaucoma by 9.5–44% [22]. Detecting glaucoma at an early stage therefore allows better disease management for patients across all populations. However, glaucoma may continue to progress even after intraocular pressure is lowered, despite it being considered a major risk factor [23].

Beyond intraocular pressure, vascular factors also play an important role in glaucoma development. Blood pressure and ocular perfusion pressure have been widely associated with glaucoma pathophysiology [24-26]. Notably, 4–7% of individuals over 40 years of age have elevated intraocular pressure (>21 mm Hg) without detectable glaucomatous damage, a condition referred to as ocular hypertension (OHT) [27]. This observation suggests that elevated IOP alone is insufficient to fully explain glaucoma onset. In addition to IOP-related factors, glaucoma risk is influenced by age, high myopia, family history, hypertension, optic disc hemorrhage, and ethnicity, with African American individuals having a substantially higher risk [28, 29]. Therefore, comprehensive eye examinations are recommended for individuals at increased risk or experiencing symptoms such as blurred vision, eye pain or redness, headaches, halos around lights, or progressive visual field loss [30, 31].

Glaucoma is commonly diagnosed using tonometry to measure intraocular pressure, visual field assessment through campimetry or perimetry, gonioscopy to examine the anterior chamber angle [32], and retinal fundus imaging to assess structural changes in the optic nerve head [33]. Among these, retinal fundus imaging (RFI) provides a cost-effective and noninvasive approach for glaucoma screening by tracking disease-related morphological changes [34]. The optic disc (OD) consists of an optic cup (OC) and a neuroretinal rim; as glaucoma progresses, increased intraocular pressure enlarges the optic cup, leading to an increased cup-to-disc ratio and a reduced rim-to-disc ratio [35, 36]. These structural measurements are used for glaucoma detection either through manual assessment or automated image analysis and machine-learning techniques [37, 38]. Recent ophthalmology research has increasingly focused on deep learning methods [39] for fundus image classification and segmentation [40], although their performance strongly depends on the availability of large, high-quality labeled datasets [41].

Automated glaucoma diagnosis has largely relied on retinal fundus images analyzed using convolutional neural networks (CNNs) [42]. Although CNNs achieve strong performance in visual feature extraction, their limited receptive fields restrict the modeling of global contextual and pathological relationships. Before the adoption of transformer-based architectures, CNNs dominated medical image analysis and computer vision applications [43]. In clinical settings, however, glaucoma assessment depends on both imaging findings and patient-specific clinical information. Integrating fundus image features with clinical biomarkers can therefore improve diagnostic accuracy, yet such multimodal approaches remain insufficiently explored. Recent reviews emphasize the need for image–clinical integration while identifying limitations in earlier studies [23, 28].

This study investigates multimodal glaucoma detection using a dataset comprising retinal fundus images and associated clinical data. Two experimental settings are considered: sample-wise and patient-wise analysis. At the sample level, glaucoma detection is evaluated using fundus images alone, clinical features alone, and a combined multimodal representation integrating both data sources. Subsequently, patient-wise glaucoma detection is performed by aggregating sample-level multimodal features at the patient level. Finally, the trained models are implemented within a web-based system to enable automated glaucoma detection and facilitate practical deployment. The key contributions of this paper are as follows:

- A multi-stage automated framework is proposed for glaucoma detection by integrating retinal fundus images with clinical data.
- Glaucoma detection is evaluated at both sample and patient levels, enabling patient-wise diagnosis through feature aggregation.
- The diagnostic performance of fundus images alone, clinical data alone, and their multimodal combination is systematically analyzed.
- A stacking-based ensemble learning approach is employed to enhance robustness and detection accuracy.
- The complete framework is implemented as an end-to-end web-based system to support automated glaucoma screening and practical clinical use.

## 2. Related Work

**Table 1** provides a reviewed summary of studies on glaucoma prediction methods, highlighting advancements and shortcomings that we aim to circumvent to enhance early detection and diagnosis.

***Table 1:*** *Summary of prior studies on glaucoma prediction methods*

| Researchers | Method Description | Datasets and Results | Advancements and Shortcomings |
|---|---|---|---|
| Veena et al. [38] | Used two separate CNNs to segment the OD and OC, calculating the CDR. | Trained on the DRISHTI-GS database; 98% accuracy for OD, 97% for OC. | Did not extend to classifying glaucoma with the segmentations. |
| George et al. [44] | Developed a 3D deep learning model with attention mechanisms, integrating three pathways for glaucoma detection. | Used a 3D-OCT cube, achieving an AUC of 0.938. | Fused outputs from multiple pathways for enhanced predictions. |
| Gheisari et al. [45] | Developed a hybrid model combining CNN with LSTM RNN to analyze spatial and temporal dynamics. | Tested on 1,810 images and 295 videos, achieving a 0.962 F-score. | Limited video data and no use of clinical biomarkers noted. |
| Shanmugam et al. [46] | Estimated CDR from fundus images using a three-stage approach including Au-Net and a random forest algorithm. | Enhanced image contrast for accurate CDR calculation and classification. | Method allows accurate classification based on CDR values. |
| Bragança et al. [47] | Developed an automated detection system using classifiers like ResNet50v2 and MobileNet. | Brazil Glaucoma dataset with 2,000 images, 90% accuracy rate achieved. | Utilized a smartphone-attached ophthalmoscope for image capture. |
| Kovalyk et al. [48] | Used neural networks, primarily ResNet50, for binary and multiclass classification. | Dataset of 244 patients, average AUC for binary classification was 0.71. | Lower AUC values in multiclass classification indicate room for improvement. |
| Sanchez-Morales et al. [49] | Developed an ensemble model combining CNNs, Capsule Networks, and Convolutional Autoencoders with K-NN. | Tested on JOINT and PAPILA datasets, high sensitivity and specificity achieved on JOINT. | Limited diagnostic accuracy due to reliance solely on image data in PAPILA. |
| Velpula et al. [50] | Proposed an XAI model using PCNNs for feature extraction and MLCs for classification. | ACRIMA dataset, 98.03% accuracy using 1,865 healthy and 1,590 glaucoma images. | High accuracy in automatic glaucoma detection highlighted. |
| Hemelings et al. [51] | Enhanced the G-RISK with labeled fundus images from significant cohorts and open-access datasets. | AUC values ranged from 0.854 to 0.988 across diverse datasets. | Broad validation with significant cohorts and open-access datasets. |
| Yi et al. [52] | Combined glaucoma-specific knowledge with deep learning for enhanced feature extraction, using U-Net and RA-ResNet. | Tested on Drishti-GS, RIM-ONE-R3, and ACRIMA, achieving high accuracies. | Validated model's efficacy across multiple datasets, could improve with patient-wise detection. |

As summarized in **Table 1**, existing glaucoma detection approaches exhibit several limitations, particularly their reliance on single data modalities and the lack of practical implementation. This study aims to address these limitations by integrating retinal fundus images with clinical data for glaucoma detection, achieving reliable diagnostic accuracy. In addition, a software-based implementation of the proposed framework is developed to demonstrate its practical applicability, which has not been considered in earlier studies.

## 3. Materials and Methods

### *3.1 Dataset Description*

The PAPILA dataset [53], collected between 2018 and 2020, contains 244 patient records and 488 fundus images, offering comprehensive information. It includes 333 entries for healthy individuals, 68 for subjects with suspicious conditions, and 87 for individuals diagnosed with glaucoma. Each record features clinical information, optic disc and cup segmentations for both eyes, color fundus images for each patient's eyes, and additional structured data.

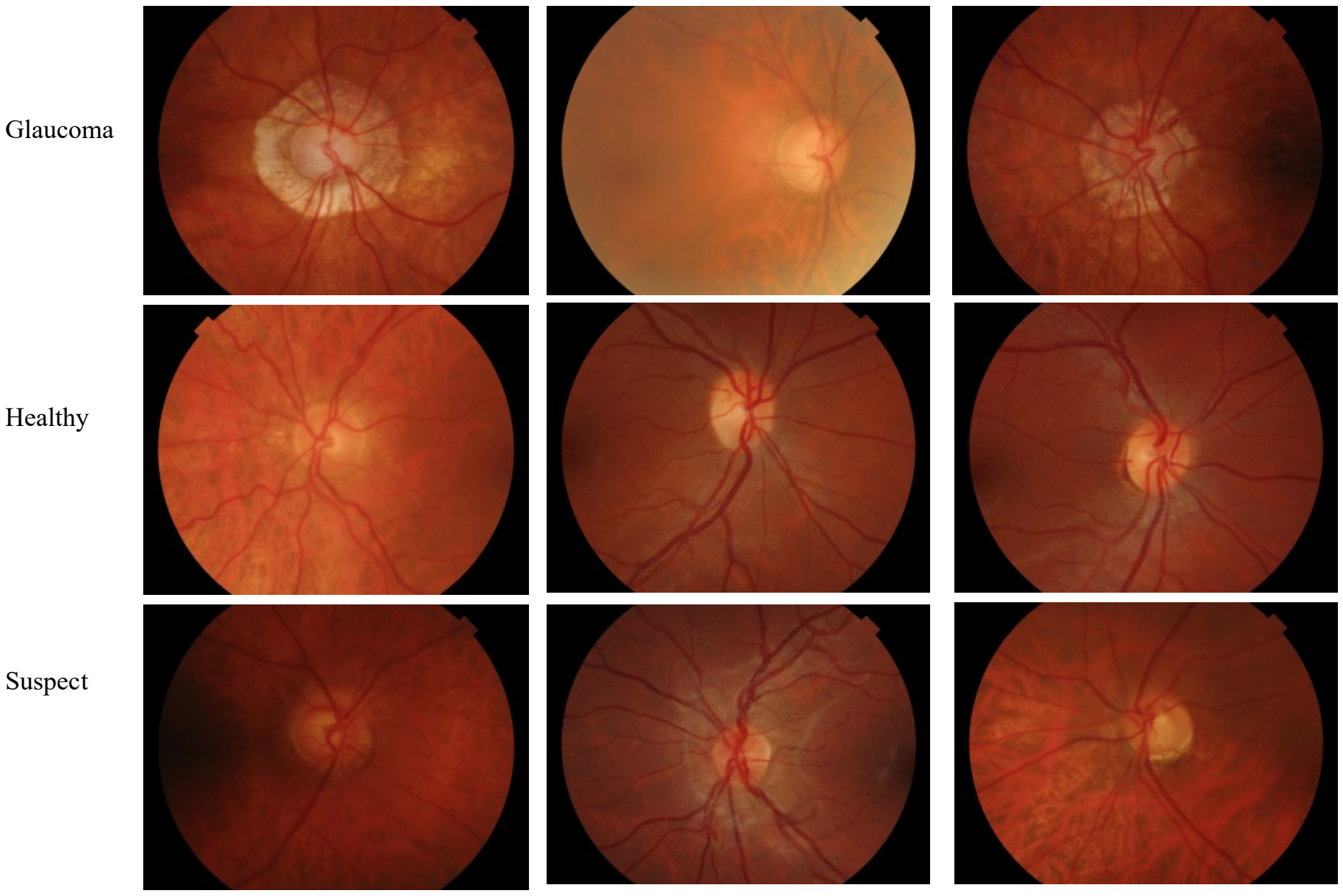

***Figure 1:*** *Sample images from the dataset*

The clinical information in the dataset includes age, gender, diagnosis, refractive error metrics (diopter 1, diopter 2, and astigmatism), phakic or pseudophakic status, intraocular pressure (measured with Pneumatic and Perkins methods), pachymetry, axial length, and visual field mean deviation (VF MD). Visual field data using the 30-2 program is specifically collected for patients with intraocular pressures exceeding 22 mmHg. **Figure 1** displays sample images from the dataset, including a glaucomatous eye, a healthy eye, and a suspect eye.

### *3.2 Methodology*

This study evaluates a stacking-based ensemble learning approach for glaucoma detection using retinal fundus images and clinical data. The two data modalities were initially processed and analyzed independently. Fundus images underwent quality screening to exclude low-quality samples, including dark and blurred images, followed by background preprocessing. To address class imbalance between glaucomatous and healthy samples, data augmentation was applied at the sample level. Deep visual features were extracted using a ViT model, after which principal component analysis (PCA) was employed to reduce feature dimensionality.

Clinical data were analyzed independently using conventional machine learning methods for binary classification. Feature engineering and selection were performed to retain informative variables and suppress noise. PCA was subsequently applied to the clinical feature space to obtain a compact representation. Stacking-based ensemble models were trained using both the original clinical features and their reduced representations.

The multimodal setting was implemented by integrating fundus-derived and clinical feature representations within a stacking-based ensemble learning framework to perform glaucoma classification at the sample level. Patient-wise glaucoma detection was then derived by aggregating sample-level predictions for each subject through grouping of images belonging to the same patient. To prevent data leakage, dataset partitioning was conducted on a subject-wise basis. Model performance was evaluated using a two-stage stacking architecture, in which three base classifiers were combined through a meta-classifier to generate the final prediction. **Figure 2** illustrates the proposed system architecture.

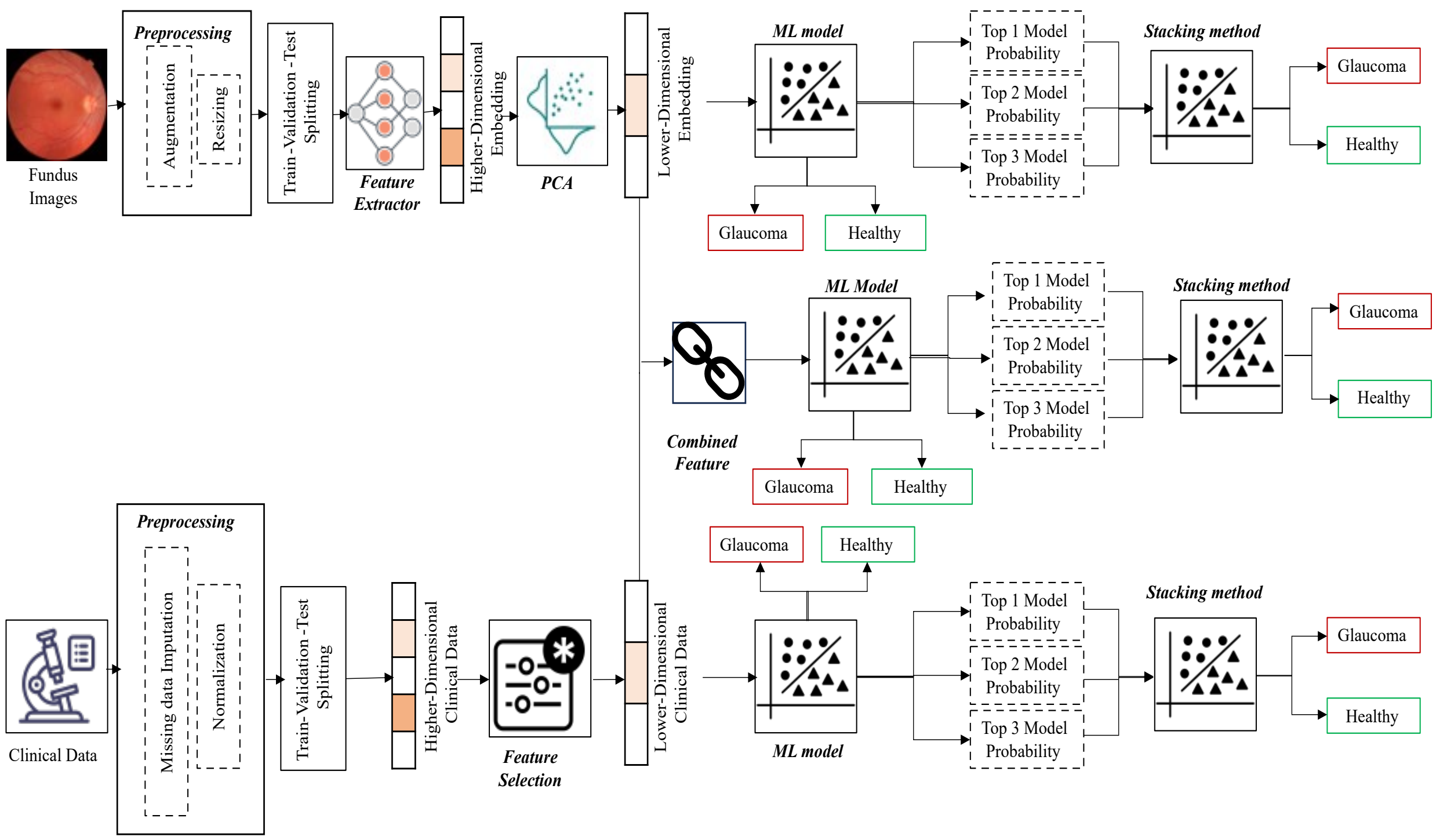

***Figure 2:** Overview of the proposed system architecture*

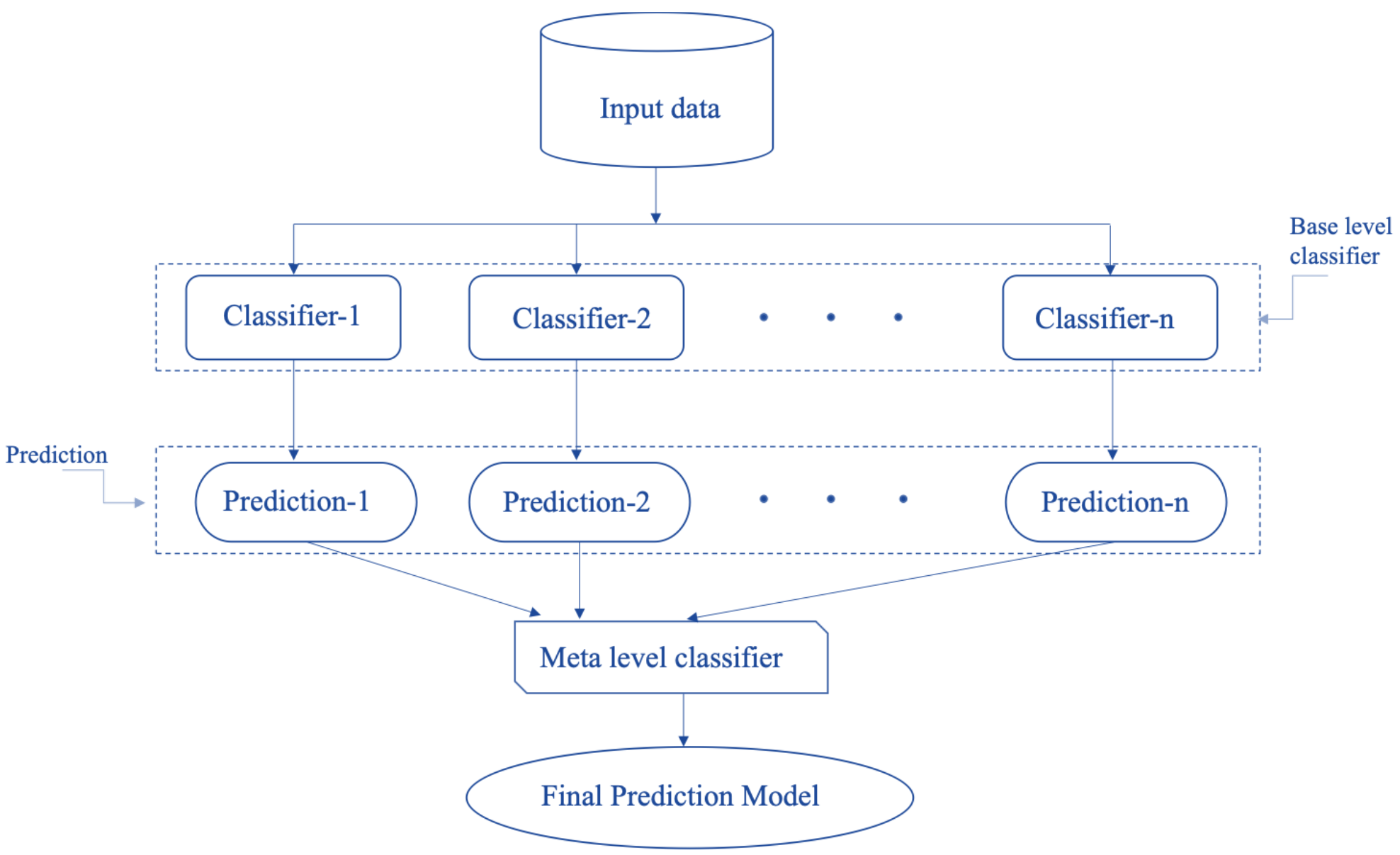

***Figure 3:** Proposed stacking model architecture*

**Figure 3** illustrates the architecture of the proposed stacking model, which combines the N-best-performing classifiers using a dataset labeled with feature vectors. Initially, n base-level machine learning classifiers generate prediction probabilities, from which a meta-learner classifier makes the final prediction based on the top-performing base learners' outputs.

### *3.3 Image Data Preprocessing*

#### *3.3.1 Data Cleaning*

The PAPILA dataset originally contained 488 fundus images, categorized based on clinical information into three groups: glaucoma, healthy, and suspect. It included 333 healthy individuals, 68 subjects with suspicious conditions, and 87 patients with glaucoma. Due to the ambiguity of the suspect class, it was excluded from further analysis. The dataset also had blurry and low-resolution images in the glaucoma category, leading to the removal of 23 problematic images to improve image clarity. Ultimately, the dataset comprised 310 records of healthy individuals and 87 records of patients with glaucoma. Error! Reference source not found. illustrates three aspects: a) the dataset before cleaning, b) examples of images retained or removed, and c) the dataset after the cleaning process.

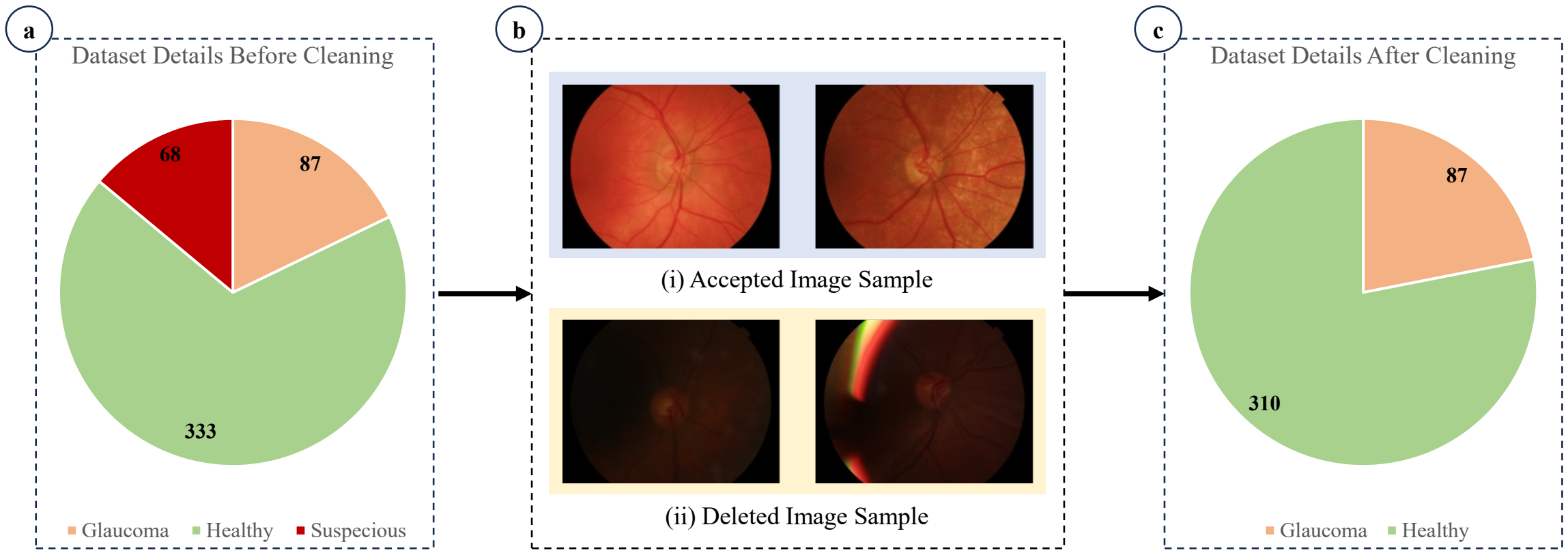

***Figure 4:** Examples of fundus images retained and excluded after quality screening*

#### *3.3.2 Data Augmentation*

In this study, image augmentation techniques such as random rotation, random horizontal flip, and padding were employed to enhance the machine learning model's ability to generalize [54]. Originally, the dataset comprised 310 images in the healthy class and 87 in the glaucoma class, presenting an imbalance. Following the augmentation, both the glaucoma and healthy groups contained an equal number of images, thereby balancing the dataset for further experiments.

### *3.4 Feature Extraction*

The Vision Transformer (ViT) model [55] was employed to extract deep feature embeddings from retinal fundus images. ViT is a transformer-based architecture that represents an image as a sequence of fixed-size patches, which are linearly embedded and processed using multi-head self-attention to model global contextual relationships. Unlike conventional CNN, ViT relies on attention mechanisms rather than convolutional operations, enabling effective capture of long-range dependencies across image regions.

ViT has demonstrated strong performance in image recognition and identification tasks across diverse visual domains [56]. In this study, the model pretrained on the ImageNet dataset was used to initialize the feature extractor. ImageNet pretraining provides robust visual representations and facilitates improved generalization, particularly in medical imaging applications with limited annotated data. The pretrained ViT implementation is publicly available and was used to extract image-level embeddings for downstream classification.

### *3.5 PCA for Dimensionality Reduction*

Principal Component Analysis (PCA) was employed to reduce the dimensionality of the feature space, transforming high-dimensional data into a lower-dimensional approximation while minimizing reconstruction error [57]. This process ensures orthogonality among all principal components, eliminating redundancy and enhancing interpretability by retaining most of the information. The new variables, termed principal components, address the eigenvalue/eigenvector solution problem. Additionally, a whitening process was applied during PCA calculation to enhance accuracy by adjusting based on data assumptions.

### *3.6 Hyperparameter Fine-tuning*

Hyperparameter tuning involves determining an ideal set of hyperparameter values for an algorithm and then using this improved algorithm to solve each given data set. This combination of hyperparameters maximizes the model's performance by minimizing a predefined loss function with fewer mistakes [58]. The hyperparameters of Logistic Regression, Linear Discriminant Analysis, and K-Nearest Neighbors (KNN) models were adjusted for better performance in this study.

### *3.7 Clinical Data Preprocessing*

#### *3.7.1 Data Imputation and Normalization*

Missing data imputation was crucial for preparing clinical data for machine learning models, as each patient had many biomarkers, though some were missing significant amounts. Instead of deleting missing entries, which could remove vital contextual information, multivariate imputation by chained equations (MICE) was employed, a technique that has shown superior performance in prior studies [59, 60]. Data normalization was also carried out using z-score normalization, where data points were standardized by subtracting the mean and dividing by the standard deviation, enhancing model effectiveness.

#### *3.7.2 Top-Ranked Features*

The feature selection was performed using the Extreme Gradient Boosting (XGB) technique, which helps in identifying traits likely to influence outcome predictions, preventing overfitting, enhancing accuracy, and reducing training time. While other feature selection methods like univariate selection, PCA, and bagged decision trees (e.g., Random Forest) are noted in the literature for their effectiveness, XGB was specifically utilized in this study to leverage the most relevant features within the dataset.

### *3.8 Experiments*

All experiments were conducted using the PyTorch package on Python 3.10.8 within a Desktop Computer Workstation operating on Windows 10 with 64 GB RAM and an i7 processor. Features were extracted using deep neural networks with PyTorch, and classification was performed using machine-learning algorithms from the Scikit-Learn library.

The study involved two types of experiments: a sample-wise classification and a patient-wise classification. In the sample-wise experiment, the dataset included seven patients with glaucoma affecting one eye; images from the affected eye were classified as glaucomatous and the other as healthy. This approach was due to the possibility of a patient having one healthy and one glaucomatous eye. In the patient-wise experiment, the seven patients with one affected eye were excluded to eliminate uncertainty, leaving 303 healthy records and 80 glaucoma records. Subjects were classified as healthy or glaucomatous based on images from both eyes.

Both experiments employed five-fold cross-validation. In each fold, 80% of the data were used for training, of which 10% was further reserved as a validation set, while the remaining 20% were used for testing. Data augmentation was implemented initially to balance the dataset, which had more healthy than glaucoma records. A five-fold weighted average of results was calculated, with the number of training and test fundus photos detailed in **Table 2**.

***Table 2:*** *Details of the dataset used for training, testing and validation*

| Database | Types of data | # of images | Training data per fold | Validation data | Augmented training data | Test data |
|---|---|---|---|---|---|---|
| **Sample-wise classification** | Glaucoma | 87 | 62 | 7 | 248 | 18 |
| | Healthy | 310 | 223 | 25 | 248 | 62 |
| **Patient-wise classification** | Glaucoma | 80* | 58 | 6 | 243 | 16 |
| | Healthy | 303* | 219 | 24 | 243 | 60 |

**NOTE: To eliminate any uncertainty, we removed the seven patients with glaucoma in one eye only for the patient-wise classification.*

### *3.9 Sample-wise classification*

#### *3.9.1 Binary Classification (Glaucoma vs Healthy) using Fundus Images*

For fundus-image–based analysis, ViT was employed as the feature extraction backbone to derive high-level visual representations from retinal images. To reduce feature redundancy and improve computational efficiency, PCA was subsequently applied, yielding a compact feature space that simplified the downstream classification task and supported improved model generalization. The resulting features were then used to evaluate multiple ML classifiers under a five-fold cross-validation protocol.

Initial experiments indicated that ML classifiers with default hyperparameter settings exhibited limited ability to reliably distinguish between healthy and glaucomatous cases. Following systematic hyperparameter optimization, the three best-performing baseline classifiers were selected as base learners. A stacking-based ensemble was subsequently constructed by integrating predictions from these three base classifiers using a meta-classifier, leading to improved discrimination between healthy and glaucoma cases compared with individual ML models.

#### *3.9.2 Binary Classification (Glaucoma vs Healthy) using Clinical Data*

In this study, clinical features were incorporated for glaucoma classification, and six features were selected from an initial set of nine variables using an XGBoost-based feature importance analysis. The selected features included VF-MD, pachymetry, axial length, Perkins, dioptre_1, and astigmatism. Using these clinical features, multiple ML classifiers were evaluated under a five-fold cross-validation protocol to identify models with strong discriminative performance between healthy and glaucomatous cases. Based on this evaluation, the three best-performing classifiers were selected as base learners. A stacking-based ensemble was subsequently constructed using these base models, and the performance of both the base classifiers and the meta-classifier was systematically assessed.

#### *3.9.3 Binary Classification (Glaucoma vs Healthy) using Fundus Images and Clinical Data*

In this experiment, features derived from fundus images and clinical data were integrated at the feature level. Image features were first extracted using ViT and subsequently concatenated with the selected clinical features to form a unified multimodal representation. Principal component analysis (PCA) was then applied to the combined feature set to reduce dimensionality and retain the most informative components. The reduced features were classified using conventional ML classifiers, followed by systematic hyperparameter tuning to optimize performance. Finally, the three best-performing classifiers were combined within a stacking-based ensemble learning framework, in which a meta-classifier integrated the base model predictions to generate the final classification.

### *3.10 Patient-wise Classification*

#### *3.10.1 Binary Classification (Glaucoma vs Healthy) using Fundus Images and Clinical Data*

For patient-wise classification, predictions were derived by aggregating sample-level outputs corresponding to each subject. Specifically, a weighted averaging scheme was applied to the prediction scores of all images associated with a given subject to obtain a final patient-level decision. This strategy accounts for the anatomical similarity between fundus images acquired from the left and right eyes, which share common retinal structures such as the optic disc, blood vessels, and surrounding tissue [61, 62]. To prevent information leakage arising from this inter-eye similarity, subject-wise data partitioning was strictly enforced. Images from different eyes of the same subject were not simultaneously included in both the training and test sets within any fold. By incorporating multiple images per subject while maintaining strict subject-level separation, this evaluation strategy reduces prediction variability and improves the reliability of patient-level classification in practical clinical settings.

### *3.11 Performance Metrics*

The efficiency of different classifiers was assessed using Accuracy (A), Recall/Sensitivity (R), Precision (P), Specificity (S), and F1-Score (F1). This study's results, based on five-fold cross-validation and the concatenation of five test-folds, depend on the entire dataset, with the metrics defined in equations (1)-(5).

$$A = \frac{TP + TN}{TP + TN + FP + FN} \tag{1}$$

$$P = \frac{TP}{TP + FP} \tag{2}$$

$$R = \frac{TP}{TP + FN} \tag{3}$$

$$F1 = 2\,\frac{Precision \times Sensitivity}{Precision + Sensitivity} \tag{4}$$

$$S = \frac{TN}{TN + FP} \tag{5}$$

The TP and TN metrics indicate the number of instances that were correctly classified as positive and negative classes respectively, while the FP and FN metrics indicate the number of instances that were incorrectly classified as positive and negative classes respectively. The model's overall accuracy can be calculated by dividing the number of TP and TN instances by the total number of instances.

## 4. Numerical Results

### *4.1 Statistical Analysis*

The clinical dataset initially comprised 11 features: age, gender, VF-MD, axial length, Perkins, diopter_1, diopter_2, phakic status, astigmatism, pachymetry, and pneumatics. Age and gender were excluded due to their well-established association with glaucoma risk, particularly in elderly populations [63, 64], and to reduce the risk of overfitting given the limited dataset size. The remaining nine clinical features were subsequently considered for analysis. The nine clinical biomarkers are: dioptre_01, dioptre_02, astigmatism, phakic/pseudophakic, pneumatic, Perkins, pachymetry, axis length, and VF_MD. Through detailed evaluation of fundus images and comprehensive review of patient eye records, these biomarkers enhance our diagnostic accuracy and patient care. **Table 3** summarizes key statistical characteristics derived from these clinical data, with gender presented both numerically and as percentages, and other variables assessed for mean, standard deviation, and missing data points. Statistical significance was determined using the Chi-square test for gender and Wilcoxon ranked tests for other variables, with p-values less than 0.05 considered significant at a 95% confidence level.

***Table 3:*** *Overview of statistical characteristics of study patients*

| Item | Glaucoma (87) | Healthy (310) | Total (397) | Statistics | P-value |
|---|---|---|---|---|---|
| **Age(years)** | | | | | |
| Number of samples | 87 (0) | 310 (0) | 397 (0) | Z=-9.51 | <0.05 |
| Mean ± STD | 70.10 ± 9.23 | 58.54 ± 12.59 | 61.08 ± 12.85 | | |
| **Gender** | | | | | |
| Female | 52 (60%) | 210 (68%) | 262 (66%) | Z = -14.01 | 0.21 |
| Male | 35 (40%) | 100 (32%) | 135 (34%) | | |
| **dioptre_1** | | | | | |
| Number of samples | 83 (4) | 296 (14) | 379 (18) | Z = 1.01 | 0.52 |
| Mean ± STD | 0.62 ± 2.16 | 0.89 ± 2.19 | 0.83 ± 2.19 | | |
| **diopter_2** | | | | | |
| Number of samples | 87 (0) | 306 (4) | 393 (4) | Z = 3.50 | 8.19 |
| Mean ± STD | -1.29 ± 0.77 | -0.98 ± 0.73 | -1.04 ± 0.75 | | |
| **astigmatism** | | | | | |
| Number of samples | 87 (0) | 306 (4) | 393 (4) | Z = -0.67 | 0.53 |
| Mean ± STD | 95.32 ± 35.39 | 92.29 ± 42.50 | 92.96 ± 41.01 | | |

| Phakic/Pseudophakic | | | | | |
|---|---|---|---|---|---|
| Number of samples | 85 (2) | 304 (6) | 389 (8) | Z = -3.46 | <0.05 |
| Mean ± STD | 0.78 ± 0.42 | 0.59 ± 0.49 | 0.63 ± 0.48 | | |
| **Pneumatic** | | | | | |
| Number of samples | 38 (49) | 296 (14) | | Z = -0.51 | 0.52 |
| Mean ± STD | 16.21 ± 3.73 | 15.89 ± 3.35 | 15.92 ± 3.39 | | |
| **Perkins** | | | | | |
| Number of samples | 69 (18) | 15 (295) | | Z = 0.37 | 0.58 |
| Mean ± STD | 17.01 ± 4.05 | 17.46 ± 4.45 | 17.09 ± 4.09 | | |
| **Pachymetry** | | | | | |
| Number of samples | 80 (7) | 309 (1) | 389 (8) | Z = -0.89 | 0.36 |
| Mean ± STD | 537.61 ± 34.16 | 533.63 ± 41.16 | 534.44 ± 39.81 | | |
| **Axial_Length** | | | | | |
| Number of samples | 82 (5) | 310 (0) | 392 (5) | Z = -1.74 | 0.204 |
| Mean ± STD | 23.71 ± 1.53 | 23.40 ± 1.09 | 23.47 ± 1.19 | | |
| **VF_MD** | | | | | |
| Number of samples | 87 (0) | 9 (301) | 96 (301) | Z = 5.42 | <0.05 |
| Mean ± STD | -7.15 ± 7.41 | -1.09 ± 2.50 | -6.58 ± 7.31 | | |

### *4.2 Best Features and their Combination Selection*

Feature importance ranking was performed using an XGBoost-based feature selection approach. Based on this analysis, VF-MD, pachymetry, axial length, Perkins, diopter_1, and astigmatism were identified as the most informative features for glaucoma classification, as illustrated in **Figure 5**.

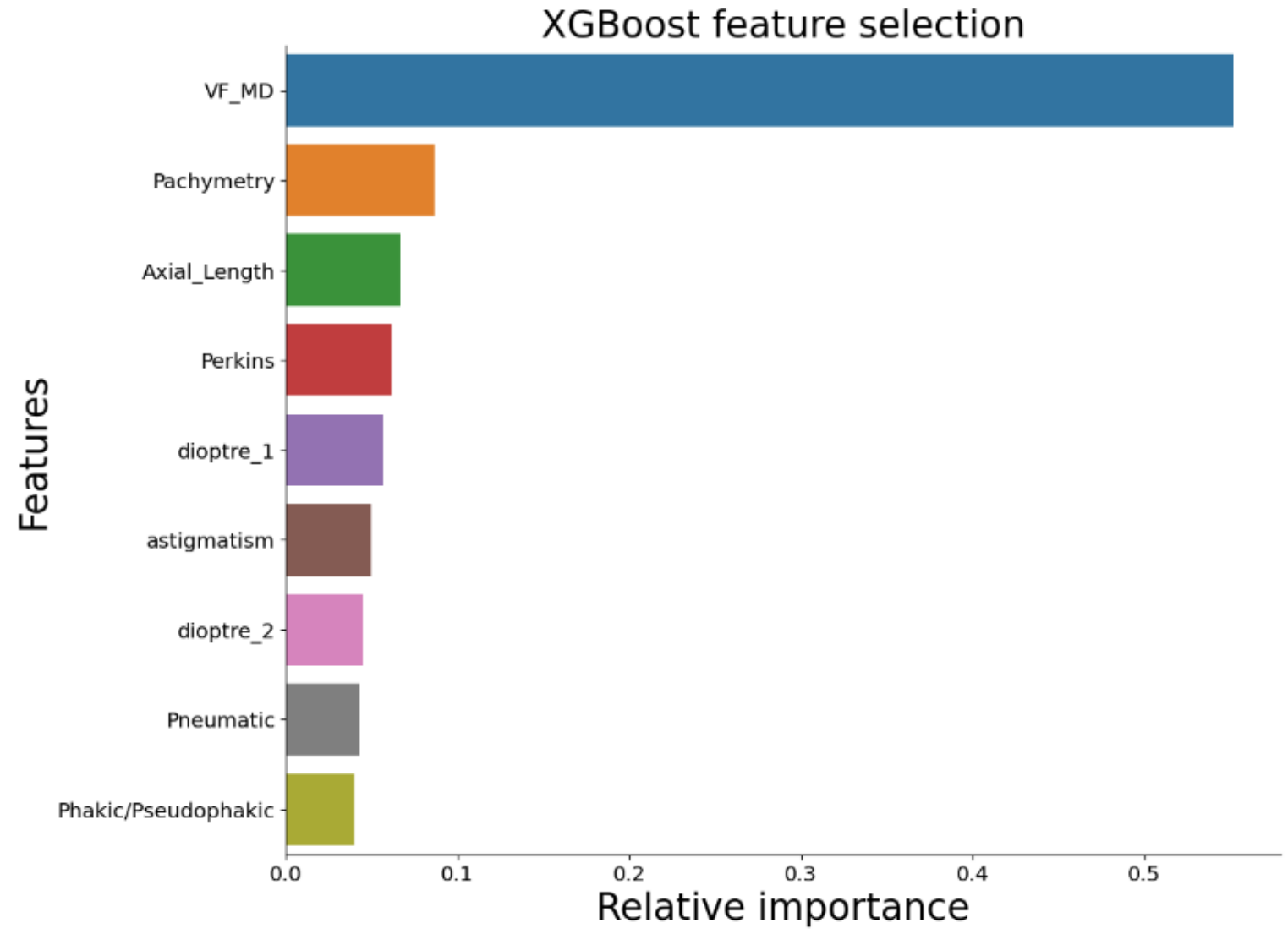

***Figure 2****: Top nine feature's rank using the XGBoost feature selection technique*

Using these selected features, multiple ML classifiers were evaluated to determine the most effective classification model. As summarized in **Table 4**, Random Forest achieved superior performance compared with other classifiers when trained on the top six features, yielding an accuracy of 96.73%, a precision of 96.83%, and an F1-score of 96.76% in the binary classification task. The selection of six features was guided by XGBoost feature importance ranking, which indicated that these variables contributed most significantly to classification performance.

***Table 4:*** *Summary of the performance metrics for Top 1 to 9 clinical features*

| Top Features | | | | | | |
|---|---|---|---|---|---|---|
| **Feature Combination** | **Model** | **Accuracy** | **Precision** | **Recall** | **Specificity** | **F1-score** |
| Top 1 feature | Random Forest | 96.47 | 96.61 | 96.47 | 95.70 | 96.51 |
| Top 2 features | | 96.47 | 96.61 | 96.47 | 95.70 | 96.51 |
| Top 3 features | | 96.47 | 96.56 | 96.48 | 94.87 | 96.50 |
| Top 4 features | | 96.22 | 96.34 | 96.22 | 94.80 | 96.26 |
| Top 5 features | | 96.22 | 96.41 | 96.22 | 95.63 | 96.27 |
| **Top 6 features** | | **96.73** | **96.83** | **96.73** | **95.77** | **96.76** |
| Top 7 features | | 95.21 | 95.42 | 95.22 | 93.69 | 95.28 |
| Top 8 features | | 95.72 | 95.84 | 95.72 | 93.83 | 95.76 |
| Top 9 features | | 96.47 | 96.56 | 96.48 | 94.87 | 96.50 |

### *4.3 Performance Analysis*

The results of sample-wise and patient-wise classification are summarized in **Table 5**, which reports performance across fundus image-based, clinical data-based, and multimodal settings.

***Table 5****: Sample-wise and Patient-wise classification performance (%) across fundus, clinical, and multimodal settings*

| **Database** | | **Classifier** | **Accuracy** | **Precision** | **Recall** | **Specificity** | **F1-score** |
|---|---|---|---|---|---|---|---|
| Fundus Images | | Linear Discriminant Analysis (LDA) | 87.91 | 87.58 | 87.91 | 87.91 | 86.93 |
| | | Random Forest (RF) | 72.80 | 84.55 | 72.79 | 72.80 | 75.12 |
| | | Logistic Regression (LR) | 86.15 | 87.52 | 86.14 | 86.15 | 86.60 |
| | | AdaBoost (AD) | 77.08 | 81.21 | 77.08 | 77.08 | 78.41 |
| | | KNeighbors (KNN) | 81.86 | 80.07 | 81.86 | 81.86 | 79.40 |
| | | **Stacking Model (LDA+LR+KNN)** | **89.92** | **89.63** | **89.92** | **89.92** | **89.70** |
| Clinical Data | | Random Forest (RF) | 96.22 | 96.34 | 96.22 | 96.22 | 96.26 |
| | | Gradient Boosting (GB) | 94.96 | 94.92 | 94.97 | 94.96 | 94.94 |
| | | AdaBoost (AD) | 95.72 | 95.68 | 95.72 | 95.72 | 95.69 |
| | | Extra Trees (ET) | 95.21 | 95.19 | 95.22 | 95.21 | 95.20 |
| | | XGBoost (XGB) | 94.96 | 95.31 | 94.96 | 94.96 | 95.06 |
| | | **Stacking Model (RF+XGB+AD)** | **96.98** | **97.11** | **96.98** | **96.98** | **97.02** |
| | Sample -wise | Random Forest (RF) | 96.47 | 96.86 | 96.48 | 96.47 | 95.55 |
| | | XGBoost (XGB) | 95.72 | 95.99 | 95.72 | 95.72 | 95.79 |
| | | AdaBoost (AD) | 96.22 | 96.24 | 96.22 | 96.22 | 96.22 |
| Fundus Images & Clinical Data | | Gradient Boosting (GB) | 94.46 | 94.42 | 94.46 | 94.46 | 94.43 |
| | | Extra Trees (ET) | 88.16 | 91.59 | 88.16 | 88.16 | 88.87 |
| | | **Stacking Model (RF+AD+XGB)** | **97.23** | **97.33** | **97.23** | **97.23** | **97.25** |
| | Patient-wise | **Random Forest (RF)** | **98.97** | **98.97** | **98.97** | **98.88** | **98.97** |

#### ***4.3.1 Performance Analysis for Sample-wise Classification***

##### *4.3.1.1 Performance Analysis using Fundus Images*

Three machine learning classifiers, logistic regression (LR), linear discriminant analysis (LDA), and K-nearest neighbors (KNN), were evaluated for glaucoma classification. LDA achieved the best results with an accuracy of 87.91%, precision of 87.58%, and an F1-score of 86.93%. While accuracy is commonly used to measure a classifier's performance, it can be misleading in imbalanced datasets, making precision and F1-score crucial for more comprehensive evaluation. To further improve results, hyperparameter tuning was applied to the top three baseline models, and a stacking model using Random Forest was constructed that combines LR, LDA, and KNN. This approach led to enhanced performance, with the stacking model achieving an accuracy of 89.92%, precision of 89.63%, and an F1-score of 89.70%.

##### *4.3.1.2 Performance Analysis using Clinical Data*

This study assessed two machine learning approaches for glaucoma classification: classical machine learning with clinical data and stacking methods using top three ML model outputs. The Random Forest (RF) classifier recorded the highest metrics with accuracy, precision, and F1-score of 96.22%, 96.34%, and 96.26%, respectively. Stacking involved Random Forest, XGBoost (XGB), and AdaBoost (AD) with a meta-learner Support Vector Machine (SVM), which outperformed the RF model, achieving 96.98% accuracy, 97.11% precision, and 97.02% F1 score. These findings suggest that SVM-based stacking of the top-performing model probabilities significantly improves glaucoma detection performance.

##### *4.3.1.3 Performance Analysis using Fundus Images and Clinical Data*

The Random Forest (RF) classifier outperforms alternative classifiers when combining fundus images and clinical characteristics with an accuracy of 96.47%, precision of 96.86%, and F1 score of 96.55%. The stacking model was developed using the top three algorithms (Random Forest, AdaBoost, and XGBoost). As a result, it achieved 97.23% accuracy, 97.33% precision, and 97.25% F1 scores, which were superior to the base models. Stacking model results demonstrate an improvement of almost 1% in overall performance when fundus images and clinical features are integrated.

#### ***4.3.2 Performance Analysis for Patient-wise Classification***

##### *4.3.2.1 Performance Analysis using Fundus Images and Clinical Data*

Using target and prediction scores from the best-performing sample-wise model, patient-wise classification predictions were generated. The Random Forest (RF) classifier excelled in patient-wise classification, achieving 98.97% in accuracy, precision, and F1 scores, demonstrating its superiority. Our study found that patient-wise classification performance is about 1% higher than sample-wise classification. **Figure 3** presents the ROC curves for sample-wise classification prior to stacking, indicating that the combined fundus image and clinical feature representation outperforms the individual modalities. The corresponding area under the curve (AUC) values for fundus image features and clinical features are 0.83 and 0.98, respectively. **Figure 7** shows the confusion matrices of image, clinical data, and multi-modal data-based classification.

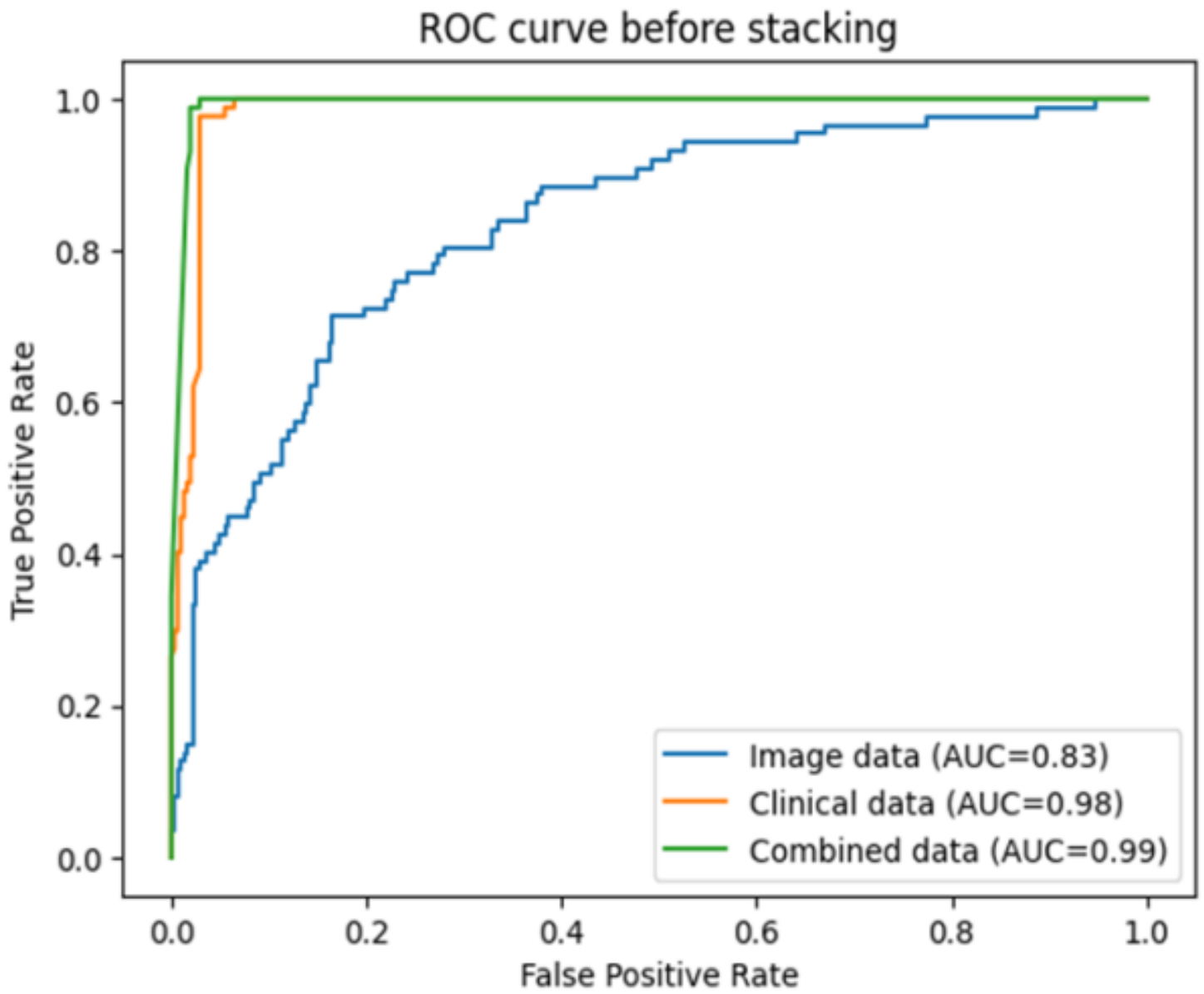

***Figure 3:*** *ROC curves for sample-wise glaucoma prediction with single and multi-modal data using the ML model before stacking*

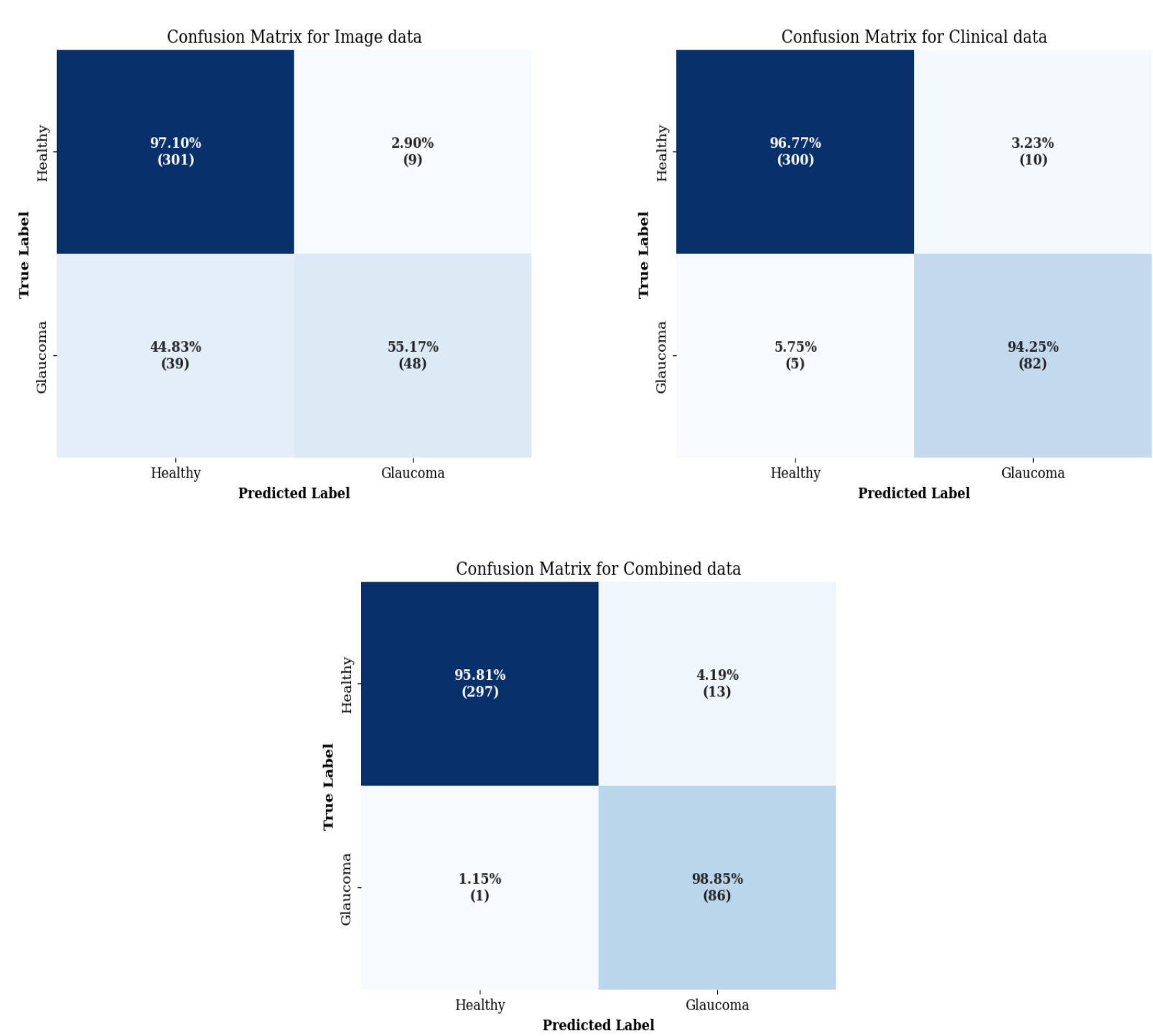

***Figure 4****: Confusion matrix for sample-wise glaucoma prediction with single and multi-modal data using the ML model before stacking*

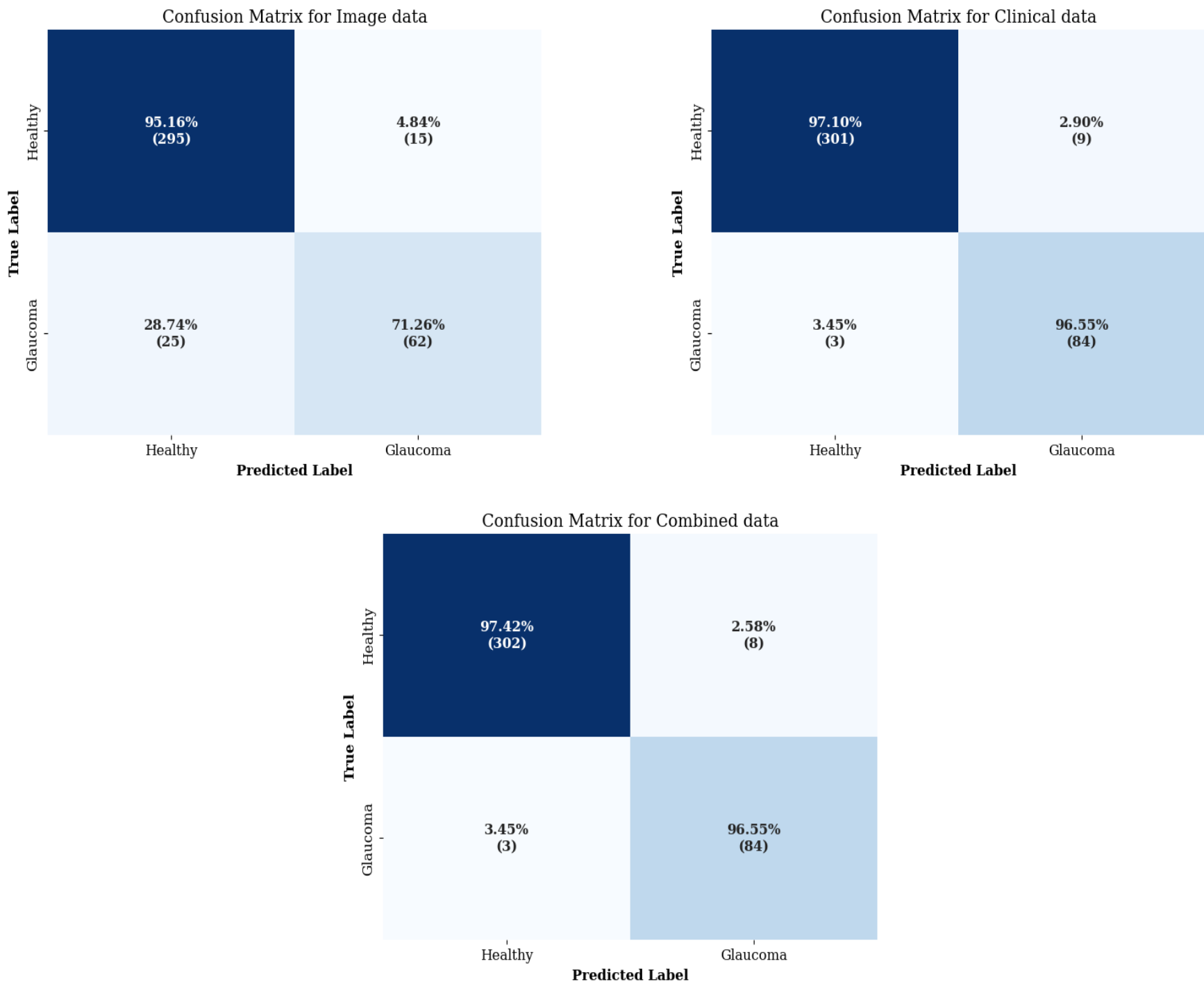

**Figure 5** displays the ROC curve for sample-wise categorization after stacking, showing significant improvements in the AUC for both clinical and imaging data. Specifically, the AUC for image data increased by 1% from 0.92 to 0.93, and when clinical and combined data are considered together, the AUC also increased by 1%. These results indicate enhanced performance across all data types—clinical, imaging, and their combination—with combined data providing superior results compared to individual data modalities.

***Figure 5****: ROC curves for sample-wise glaucoma prediction with single and multi-modal data using the ML model after applying stacking*

***Figure 6:*** *Confusion matrix for sample-wise glaucoma prediction with single and multi-modal data using*

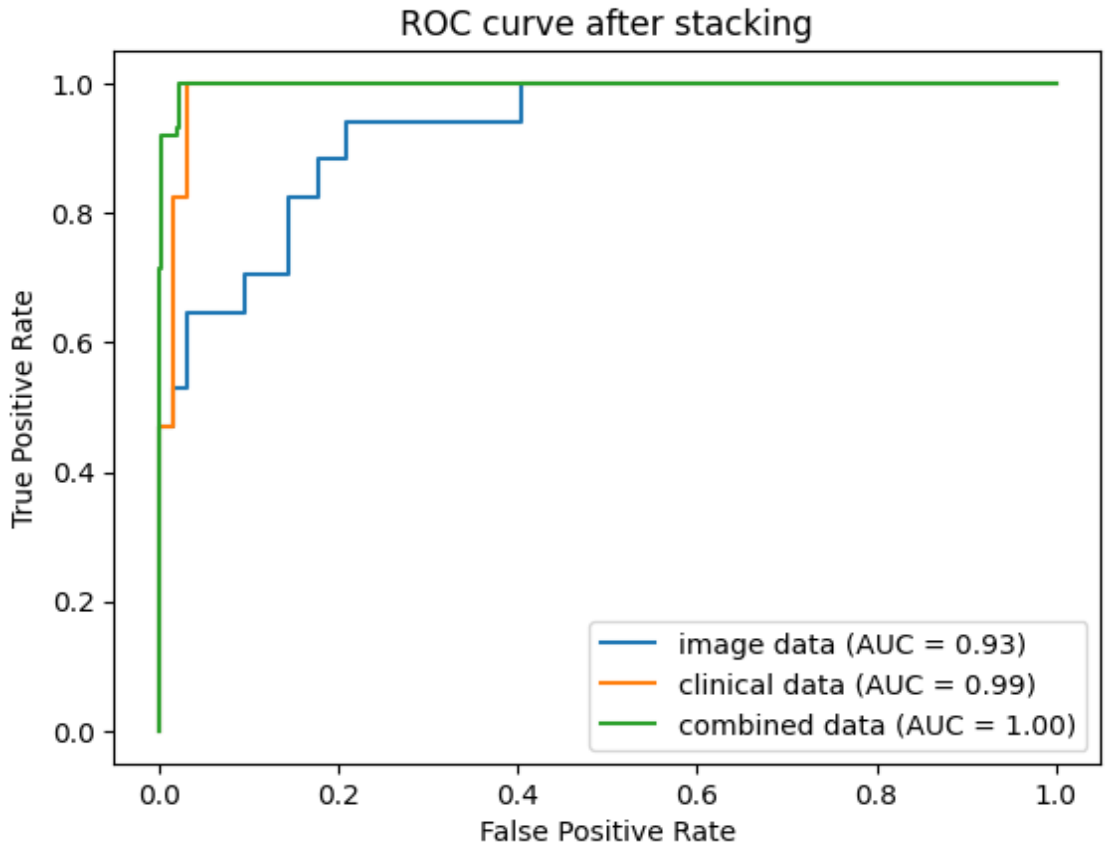

*the ML model after applying stacking*

The confusion matrix for sample-wise classification with stacking can be seen in **Error! Reference source not found.**. According to the analysis of data, glaucoma cases are reduced improperly labeled. In particular, the number of mislabeled instances for image data decreased from 39 to 25. Similarly, clinical data mislabeling decreased from 5 to 3. As a result of the improvement, the labeling process is more accurate; stacking mitigates the severity of misclassifications as compared to not stacking.

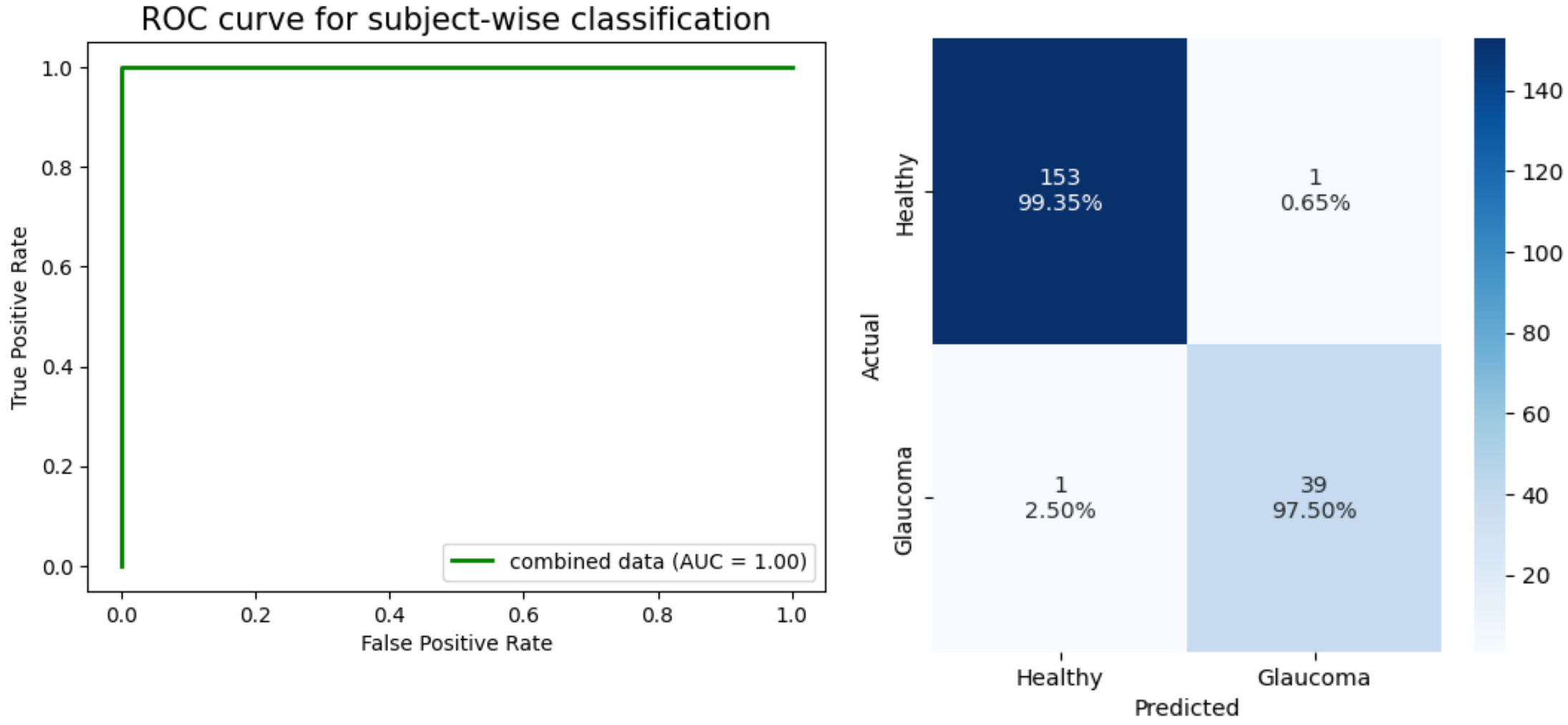

***Figure 7****: ROC curve and confusion matrix for patient-wise glaucoma prediction multi-modal data using the ML model*

**Figure 7** presents the ROC curve and confusion matrix for patient-wise classification of multi-model data, where the combined data AUC scores a perfect 1, highlighting exceptional discriminative ability. The patient-wise dataset consists of 383 records, with 303 classified as healthy and 80 as glaucoma. After group-by-group analysis, the dataset reduced to 194 records—154 healthy and 40 glaucoma. Patient-wise classification shows a very low misclassification rate of 1 in the confusion matrix, significantly improving the reliability and accuracy of glaucoma diagnosis compared to sample-wise classification.

**5. Web Application with Backend Server**

During this research, we developed an online tool called Glaucoma Prediction that allows physicians to monitor patients' eye health effectively. This system uses three prediction techniques and offers assessments based on fundus images, clinical data, or both, which are typically in PNG or JPEG formats. Users upload their data to the server where it is preprocessed and analyzed to determine the presence of glaucoma. Results are displayed on the screen and recorded in an SQLAlchemy database for accurate recordkeeping and traceability. This tool not only facilitates early detection and intervention for patients but also reduces the healthcare system's burden. **Figure 11** displays the website's homepage.

***Figure 8:*** *Homepage of Glaucoma Prediction website*

Below are the characteristics of the 'Glaucoma Prediction' system:

1. **Predict Glaucoma based on fundus images:**
   Users upload retinal fundus images during registration. Upon clicking the predict button, the server analyzes these images to assess the likelihood of glaucoma.

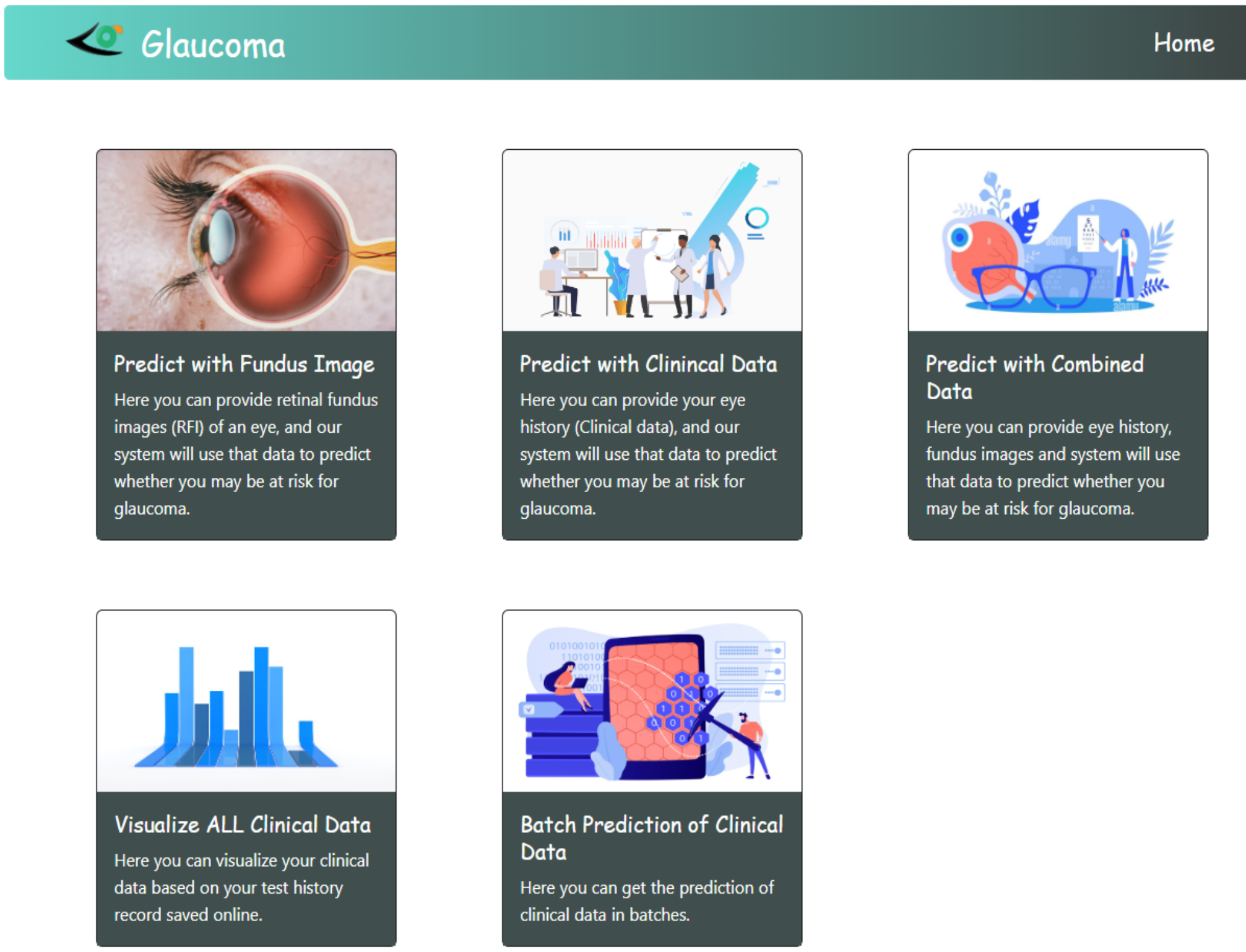

2. **Predict Glaucoma based on clinical data:**
   The system evaluates the user's risk of glaucoma using clinical data such as VF-MD, pachymetry, axial length, Perkins, diopter-1, and astigmatism, applying advanced machine learning models.
3. **Predict Glaucoma from Combined Data (fundus image and clinical data):**
   Users must provide both retinal fundus images and ocular history. The system integrates these data types to more accurately determine the user's health status, enhancing reliability over methods using singular data types.
4. **Clinical records visualization:**

The system offers visualizations of eye metrics to aid healthcare professionals in assessing ocular health. Interactive bar plots show changes over time, helping identify trends and anomalies for informed decision-making.

5. **Batch processing of clinical data:**
Ideal for hospitals managing large patient volumes, this feature allows for efficient risk assessment of glaucoma via a CSV file upload, automating the prediction process for multiple patients simultaneously.

## 6. Discussion

In this study, we proposed an automated glaucoma detection framework using a multimodal learning approach by integrating fundus images with clinical data for glaucoma detection. The resulting multimodal framework achieved an F1-score of 97.25%, compared with 89.70% for fundus-based models and 97.02% for models using clinical features alone. A similar trend was observed for the AUC metric, with values of 0.93 for fundus imaging, 0.99 for clinical data, and 1.00 for the combined modality. The stacking-based ensemble produced higher classification performance than models trained on individual data sources. In addition, patient-wise classification consistently outperformed sample-wise classification, with both accuracy and F1-score reaching 98.97% and an AUC of 1.00, indicating that aggregating multiple images at the subject level provides more reliable predictions than single-image-based inference.

Most prior studies on automated glaucoma detection have primarily relied on segmentation-based analysis of retinal fundus images, with a subset employing convolutional neural networks or ensemble classifiers for disease classification. In contrast, the proposed approach integrates fundus images with clinical features within a unified multimodal framework, as summarized in **Table 6**, and demonstrates improved diagnostic performance relative to existing methods. For instance, Bragança et al. [47] and Sanchez-Morales et al. [49] employed different ensemble strategies but reported comparatively lower AUC values. Hemelings et al. [51] achieved the highest reported AUC of 0.88 on the PAPILA dataset, which remains approximately 13% lower than the AUC obtained by the proposed model. This difference reflects the effectiveness of combining visual representations derived from a Vision Transformer–based feature extractor with complementary clinical variables. In addition, fine-tuning of the classification models and the use of a stacking-based ensemble further contributed to the observed performance gains.

*Table 6: Comparison of the study with contemporary state-of-the-art literature*

| Reference | Method | Dataset | Results |
|---|---|---|---|
| **Bragança et al. [47]** | ensemble model incorporated with Resnet50v2, Mobilenet, Densenet, InceptionV3, and Resnet101 | Brazil Glaucoma (BrG) | AUC: 0.965<br>Accuracy: 90% |
| **Sánchez-Morales et al. [49]** | Ensemble model based on the KNN algorithm | JOINT And PAPILA | JOINT dataset:<br>AUC: 0.98<br>PAPILA dataset:<br>AUC: 0.78<br>Specificity: 74% |
| **Hemelings et al. [51]** | Improved glaucoma referral regression network (G-RISK) model | Two population cohorts (BMES and GHS), 11 datasets (AIROGS, ORIGA, REFUGE1, | AUC (BMES and GHS): 0.976 and 0.984 |
| | | LAG, ODIR, REFUGE2, GAMMA, RIM-ONEr3, RIM-ONE DL, ACRIMA, and PAPILA). | AUC (11 datasets): 0.854 ~ 0.988<br><br>AUC (PAPILA): 0.88 |
| **Kovalyk et al. [48]** | CNN | PAPILA | AUC: 0.71 |
| **Proposed study** | Multimodal (fundus + clinical) data, ViT-based feature extraction, stacking-based ML ensemble classification, and web-based system | PAPILA | **Image-wise:**<br>AUC: 1<br>Accuracy & Specificity: 97.23%<br>**Subject-wise:**<br>AUC: 1<br>Accuracy: 98.97%<br>Specificity: 98.88% |

In our study, we enhanced glaucoma detection by combining features from both fundus images and clinical data using a multi-model approach, demonstrating improved results over existing methods that rely solely on one type of data. We employed a ViT model, which, unlike CNNs, adapts to changes in input and captures both local and global features, allowing for faster and higher-quality data processing. We optimized the model's parameters to maximize performance and integrated various base models into a stacking-based meta-learner, enhancing reliability and robustness. Patient-based classification, utilizing multiple images per patient, showed higher performance compared to sample-based methods, suggesting potential for further improvement with more extensive datasets. To facilitate practical adoption, the proposed framework has been implemented as a dedicated software platform that integrates multimodal prediction within a unified interface; to the best of our knowledge, such an end-to-end implementation has not been reported in prior glaucoma detection studies.

While the proposed multimodal framework demonstrates strong performance, the present study is conducted within a defined experimental scope that naturally introduces certain constraints. The dataset size, although sufficient for patient-based evaluation, remains limited in terms of population diversity and acquisition variability, which may influence generalization across different clinical settings. In addition, imaging device heterogeneity and demographic factors were not explicitly modeled, as the primary focus of this work was on evaluating the effectiveness of multimodal feature integration rather than domain adaptation. The stacking-based ensemble, while improving classification stability, also increases model complexity, which may affect computational efficiency in large-scale or real-time deployments. Furthermore, the framework currently relies on supervised learning with structured clinical variables, and its performance may be influenced by the availability and completeness of such data in routine practice.

Future work may extend the validation of the proposed framework to larger, multi-center datasets encompassing diverse imaging protocols and patient populations. Strategies aimed at reducing computational complexity and the use of semi-supervised or self-supervised learning paradigms may also be explored to improve robustness in settings with limited annotations. In addition, incorporating clinically oriented interpretability components and applying the framework to other ophthalmic conditions may help assess its broader applicability.

## 7. Conclusion

This work demonstrates that integrating fundus images with clinical variables within a stacking-based ensemble framework enables accurate glaucoma detection, achieving up to 98.97% patient-wise performance and an AUC of 1.00. Relative to sample-wise evaluation and non-stacking configurations, the proposed patient-wise stacking strategy yields an improvement of approximately 1%, indicating more consistent and reliable predictions when both multi-sample aggregation and ensemble fusion are employed. To support practical use, the method has been implemented within a dedicated software platform designed to facilitate efficient and accessible glaucoma screening. Together, these findings suggest that multimodal modeling, coupled with deployable tools, can contribute to more robust computer-assisted glaucoma assessment.

**Declarations:**

- **Competing interests:** None declared
- **Ethical approval:** Not required
- **Institutional Review Board Statement:** Not applicable
- **Informed Consent Statement:** Not applicable
- **Data Availability Statement:** The processed dataset used in this study can be made available upon a reasonable request to the corresponding author.